\documentclass{article}

\usepackage{arxiv}

\usepackage[utf8]{inputenc} 
\usepackage[T1]{fontenc}    
\usepackage{hyperref}       
\usepackage{url}            
\usepackage{booktabs}       
\usepackage{amsfonts}       
\usepackage{nicefrac}       
\usepackage{microtype}      
\usepackage{tcolorbox}
\usepackage{graphicx}
\usepackage[numbers]{natbib}
\usepackage{xcolor}
\usepackage{tikz}
\usepackage{amsmath}
\usepackage{xspace}
\usepackage{listings}
\newcommand{\wikitask}{SynthWiki\xspace}

\begin{document}

\title{Attention Sorting Combats Recency Bias In Long Context Language Models}


\author{Alex Peysakhovich\thanks{alex.peys@gmail.com. Currently at Sutter Hill Ventures}\\
	 \And
	Adam Lerer\thanks{adam.lerer@gmail.com. Currently at Google DeepMind} \\
}


\renewcommand{\shorttitle}{Attention Sorting Combats Recency Bias}

\maketitle

\begin{abstract}
Current language models often fail to incorporate long contexts efficiently during generation. We show that a major contributor to this issue are attention priors that are likely learned during pre-training: relevant information located earlier in context is attended to less on average. Yet even when models fail to use the information from a relevant document in their response, they still pay preferential attention to that document compared to an irrelevant document at the same position. We leverage this fact to introduce ``attention sorting'': perform one step of decoding, sort documents by the attention they receive (highest attention going last), repeat the process, generate the answer with the newly sorted context. We find that attention sorting improves performance of long context models. Our findings highlight some challenges in using off-the-shelf language models for retrieval augmented generation.
\end{abstract}

\section{Introduction}


Recent techniques have allowed large language models (LLMs) to operate on extremely large context corresponding to dozens of pages of text \citep{beltagy2020longformer,press2021train,touvron2023llama,chen2023extending,peng2023yarn}. 
These long contexts can be populated with user history, documentation, detailed model instructions, or few-shot exemplars, allowing the model to dynamically extract and integrate diverse information to complete its task \citep{schick2023toolformer,rubin2023long,shi2023replug}. We will refer to such tasks as context augmented generation. A major use case in practice is retrieval augmented generation (RAG) where in addition to receiving a query, the model receives additional documents in context from a retriever system \citep{lewis2020retrieval,ram2023context}. RAG is ubiquitous in applications such as question answering, code completion, and summarization. 

However, whether transformers can actually use long context effectively is a major area of study \citep{tay2020long,sun2021long,anadim2023thelittleretrievaltest,liu2023lost}. For example \cite{liu2023lost} finds that LLMs are worse in a RAG task when the relevant document is placed in the middle of the context. The authors mention that ``observing a serial-position-like effect in LLMs is perhaps surprising, since the self-attention mechanisms underlying Transformer LLMs is technically equally capable of retrieving any token from their contexts.''

A likely culprit for this phenomenon is a mismatch between the task LLMs are trained on and context-augmented generation tasks. Among the documents typically used to pre-train LLMs such as web pages, books, articles and code, the most informative tokens for predicting a particular token are typically the most recent ones. During pre-training, this induces a learned bias to attend to recent tokens. In addition, the rotary positional embedding (RoPE) scheme used in the open source models we investigate has an inductive bias towards reduced attention at long distances \citep{su2021roformer} that may make it even easier for these models to learn to attend preferentially to recent tokens. Extreme recency bias is not a good prior for context augmented generation tasks where far away tokens may, in fact, contain very relevant information.

We study this hypothesis in the open-domain question-answering \citep{lewis2020retrieval} environment using a `one-relevant-document-many-irrelevant-distractors' task (distractor QA). Distractor QA is a simple testbed for examining LMs' ability to make use of their context. Our task documents are one-paragraph biographies of fictional individuals generated by a language model. The model is asked simple questions that can be readily answered from the document. The distractors are biographies of other fictional individuals. Our task maintains the structure of a typical QA task but, because all data is generated, incorporates a useful feature that models cannot use their prior knowledge to help with the task. We call this the \wikitask task.

We study the performance of four small open source long-context models and three proprietary long-context models released by OpenAI and Anthropic. 

The open source models we study are \href{https://huggingface.co/togethercomputer/LLaMA-2-7B-32K}{TogetherComputer's Llama-2-7B-32k}, \href{https://huggingface.co/togethercomputer/LLaMA-2-7B-32K-Instruct}{Llama-2-7B-32k-Instruct}, YaRN Llama-2-7B-64k \citep{peng2023yarn}, and WizardLM-tuned Code-Llama-7B \citep{luo2023wizardcoder}. They are all based on Llama-2 which pre-trained with a base context window of size $4096$, but increase the context window via variations of positional encoding extension and fine tuning. TogetherComputer's models are of particular interest as they are fine tuned not only on long context next token prediction but explicitly on long context QA. 

We find that increasing the number of distractors degrades model performance across both small open source LLMs and larger proprietary ones. The decrease varies across models with the smallest degradation in Claude-1-Instant and Claude-2, a larger degradation in GPT-3.5 and TogetherLlama, and catastrophic degradation in the remaining open source models (Figure \ref{fig:synthwiki_acc}).

Unlike the API-based proprietary models, the open source models give us access to the model internals. This allows us to dive deeper into the cause of the degradation. We find that the closer a relevant document is placed to the generated response in context, the more attention weight it receives. However, we still see that on average true documents receive much more attention than distractors \textit{at the same position in context}. This gives rise to the idea of ``attention sorting'': run one decode step and look at per-token attention weights averaged across layers and heads, sort the documents by these attention scores, optionally repeat this process,\footnote{Intuitively, multiple passes are helpful because in each pass, the relevant document(s) will tend to have a higher attention score than distractors at the same position, and will preferentially move towards the end of the context. Usually by two passes true documents are within the `useable' part of the model's context.} and finally run model generation with this newly sorted context. 

We find that attention sorting greatly improves upon model QA accuracy in our task for all 3 models we study (Figure \ref{fig:main_fig}), and allows TogetherComputer's Llama to match or exceed the performance of the best proprietary models on our QA task. We hypothesize the higher accuracy and better re-sorting performance of TogetherLlama is because that particular model has been specifically tuned for long context QA while WizardCoder and YaRN were not. Nevertheless, we find that attention sorting still substantially helps the latter models - particularly with repeated application - more than doubling their accuracy at 30k context length. 

\section{\wikitask: A Synthetic Dataset for Long-Context Extractive QA}
We are interested in studying context-augmented generation where an LLM receives a question $q$ and a set of document $D$ and outputs an answer $a(q, D).$ This is sometimes called retrieval augmented generation or extractive open domain QA.

We will study the simplest case where the model needs to extract just a single piece of relevant information. Here $D$ contains a document $d^*$ which contains the answer to the question and other documents $D_{distract} = d_1, \dots, d_k$ which do not contain any information related to the question. We will study how increasing the size of $D_{distract}$ affects LLM performance. 

A known issue with LLMs is that they tend to use knowledge priors from training even when the instruction asks the model to only use information from the context \citep{wei2023larger}. In order to study long-context question answering in isolation, we generate a synthetic dataset of documents that minimizes knowledge contamination from pre-training.

We generate our dataset using a madlib procedure: we first use GPT3.5 to generate combinations of nationalities, jobs, and names for a fictional person. We have a set of $20$ question types (examples: ``where did $X$ go to school?'', ``what year was $X$ born?'' see Appendix \ref{app:synthwiki_construction} for full list). We sample $2$ such questions, we then use GPT4 to general a short ``one paragraph Wikipedia article'' summary for this person which contains the answers to these two randomly sampled questions. We have some cleaning logic to remove bad questions or poorly formatted replies. We call this dataset SynthWiki and plan to release code to generate new documents to avoid future contamination. Our task can be thought of as a natural language version of the `Little Retrieval Test' \citep{anadim2023thelittleretrievaltest} where models are, essentially, asked to retrieve a value from a long key, value list.

For the experiments reported here, we generate $500$ random SynthWiki entries with $2$ questions per document. After some minor cleanup we are left with $990$ entries in our dataset. An example biography is below:

\begin{quote}
Amalia Varela is a renowned Argentine economist, best known for her significant contributions to fiscal and monetary policies in Argentina. Born in the bustling city of Buenos Aires, Varela established herself as a passionate advocate for economic reform and sustainable development in her home country. She has served as an adviser to various governmental and non-governmental organizations, shaping Argentina's economic landscape through her innovative theories and rigorous research. A true football enthusiast at heart, Varela is also known for her unwavering support of Racing Club, a well-respected football team based in her birth city. Throughout her illustrious career, Varela’s fusion of economics and devotion to her favorite sports team have solidified her status as a transformative figure in Argentine society.
\end{quote}

This passage would be paired with questions like: ``In which city was Amalia Varela born?'' or ``What is the name of Amalia Varela's favorite sports team?.'' The average document length is $165$ tokens when using the Llama/CodeLlama tokenizer. The format of the QA prompts follows closely to the model pages on HuggingFace and is shown in Appendix \ref{app:synthwiki_prompt}.

We use SynthWiki to generate $K$ examples of $(q_i, a_i, d_i).$ To construct the set $D_{distractor}$ we set a maximum context size $\hat{T}$ and sample $d_j$ for $j \neq i$ without replacement to fill the context size. Each of these documents are irrelevant distractors for the true document. We randomize the order of $D$ when we present it to the model.

\section{Performance of Long-Context LLMs on \wikitask}
\begin{figure}[ht!]
    \centering
    \includegraphics[scale=.7]{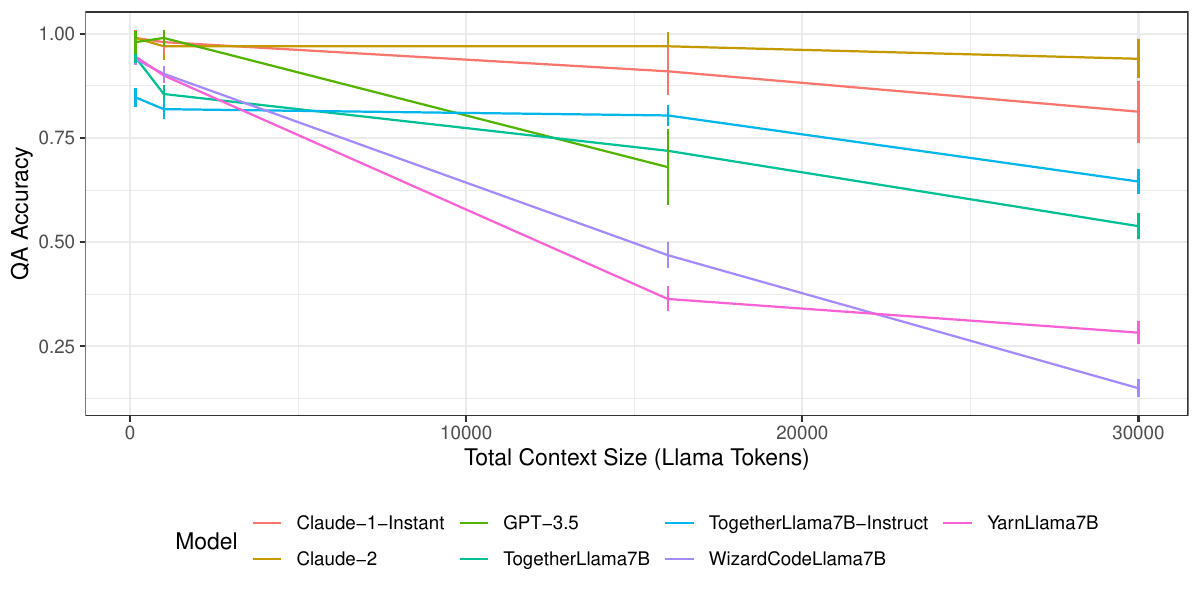}
    \caption{Performance of all long-context models that we study on question answering degrades when the relevant information is embedded in a long context of irrelevant distractor text.}
    \label{fig:synthwiki_acc}
\end{figure}

We study the following set of models:
\begin{enumerate} 
\item \href{https://huggingface.co/togethercomputer/LLaMA-2-7B-32K}{togethercomputer's Llama-2-7B-32k} model extends the base Llama-2 \citep{touvron2023llama} context via position interpolation of the original models. In addition, this model is specifically fine tuned on long context summarization and QA tasks.
\item \href{https://huggingface.co/togethercomputer/LLaMA-2-7B-32K-Instruct}{togethercomputer's Llama-2-7B-32k-Instruct} performs additional fine tuning on a different recipe of instruction following, multi-turn chat, book summarization, and more long context QA.
\item \href{https://huggingface.co/WizardLM/WizardCoder-Python-7B-V1.0}{Wizard Code Llama-2 7B} is an instruction tuned version of Code LLama 2 \citep{luo2023wizardcoder}. The baseline Code Llama 2 already comes with an extended context up to 100k which is extended via long-context next token prediction. We choose the Wizard version as it comes with explicit instruct tuning.
\item \href{https://huggingface.co/NousResearch/Yarn-Llama-2-7b-64k}{YaRN Llama-2-7b-64k} \citep{peng2023yarn} uses a novel method for extending context combined with next-token prediction based fine tuning. We note that this model is \textit{not} instruction tuned. At the time of this writing, the authors are not aware of any instruct-tuned versions of this model. We will update the paper as soon as they become available.
\end{enumerate}

We also look at some prioprietary models available through APIs:
\begin{enumerate}
\item GPT-3.5-turbo-16k is a 16k context version of OpenAI's GPT 3.5. As of this writing, no details on how 16k context is achieved are available.
\item Claude-2 and Claude-1-Instant are 100k context models from Anthropic. Here, again, details on architecture are unavailable.
\item At the time of this writing, neither author had access to GPT-4-32k. We plan to update this paper when we are able to access the API.
\end{enumerate}

We use the production versions of these models as of September 2023.

We first see how accuracy at SynthWiki changes as we change the total context size by adding distractor documents. For the open source models we evaluate on all $990$ questions (445 distinct documents), while for the API-based models we evaluate a sample of $100$ random questions.

We report the context lengths for all models in terms of the Llama tokenizer for a fair comparison between models. Since the GPT and Claude tokenizers have different vocabulary size than the Llama tokenizer, their contexts will, on average, correspond to fewer tokens for the same text input.

We report results for context lengths of $1000$, $16,000$, and $30,000$ (Llama) tokens. We fill the contexts by randomly sampling distractor documents until the next sample would cross the maximum context we allow. The document order is shuffled so the true document is in a random position. This gives us on average 5.5, 95, and 180 total documents, respectively.

We remove quotations from model responses, and count a model response as correct if it contains an exact string match to the true answer. Manual inspection of random samples finds that 
approximately 2\% of correct answers are classified as incorrect due to variation in possible ways to answer the question. 

We see that in all models accuracy decreases with context length (Figure \ref{fig:synthwiki_acc}). All trends look as expected, except we see that Together-Llama-Instruct seems to be less accurate in short contexts than the other 7B models but does significantly better in longer contexts. It is unclear to us why this is.

\begin{figure}[h!]
    \centering
    \includegraphics[scale=.75]{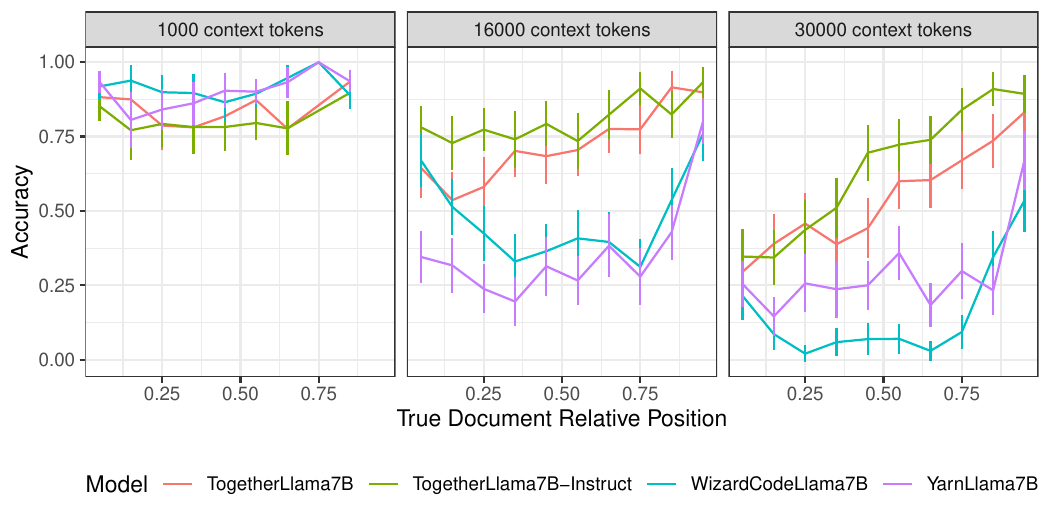}
    \caption{QA accuracy on \wikitask as a function of the position of the relevant document in the context. We see a replication of the `lost in the middle'\citep{liu2023lost} effect on this dataset in which accuracy is lower when the relevant information is in the middle of a long context. The recency (information toward the end of the context) effect seems to be quite general across models and context lengths. However, the primacy (first documents) effect seems to be less general.}
    \label{fig:acc_lost}
\end{figure}

In Figure \ref{fig:acc_lost} we plot accuracy as a function of the position of the true document in the context. We see a general replication of the `lost in the middle' results \citep{liu2023lost} which finds that LLMs prefer information in the beginning of the contexts (primacy bias) and end (recency bias).  However, we see that as context gets longer, and in  togetherLlama in particular, the primacy bias towards early documents becomes smaller and we see mostly the recency bias.

\begin{figure}[h!]
    \centering
    \includegraphics[scale=.75]{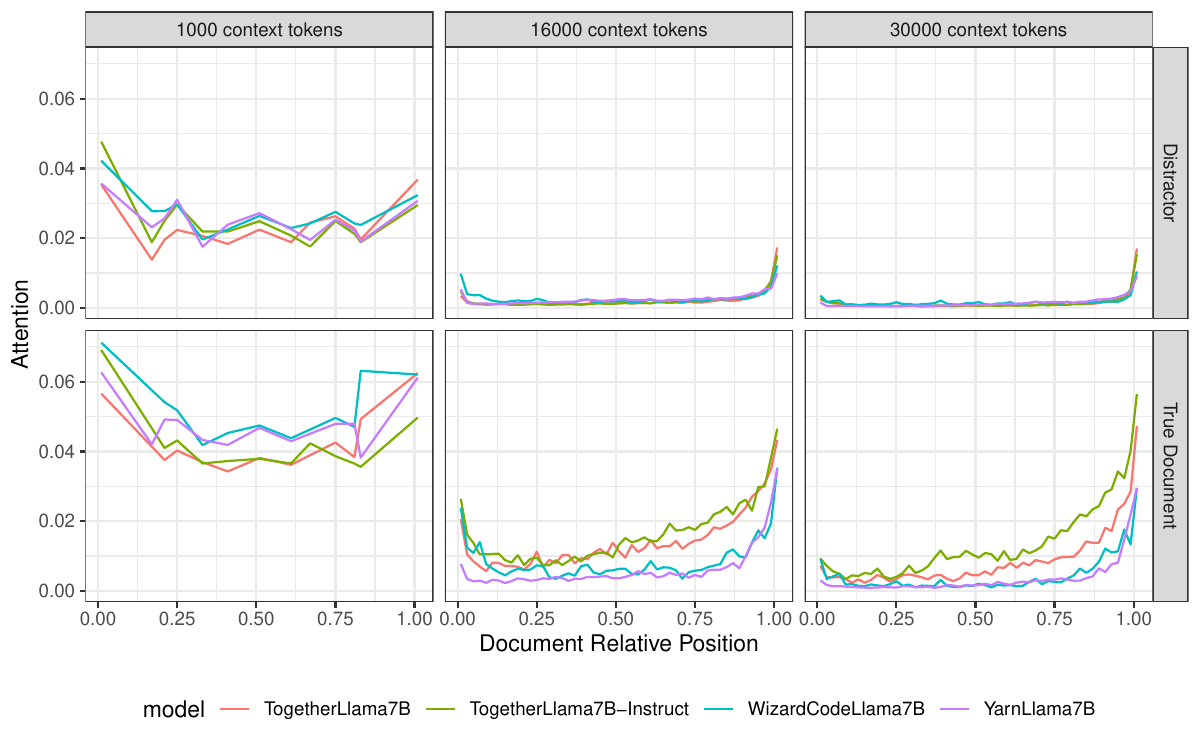}
    \caption{Average attention weight by source token position for different context lengths, averaged over all layers and attention heads. The attention weights are only computed for the first generated response token. At long context lengths, all three models show a strong bias towards attending to the most recent tokens, as well as a weaker bias towards the initial tokens. All models also attend much more strongly to relevant documents than distractor documents.}
    \label{fig:attn}
\end{figure}

In Figure \ref{fig:attn} we plot model attentions during the first step of generation. We sum the post-softmax model attentions across heads and document tokens to get an attention score for each document. We plot the attention scores for each model based on document position and whether the document is the true document for the given question. 

We again see a strong recency bias in average attention for longer contexts. But we also observe that the document containing the answer generally has higher much higher attention than distractors at the same position in context. 

\section{Re-Sorting Documents by Attention}
\begin{figure}[h!]
    \centering
    \includegraphics[scale=.25]{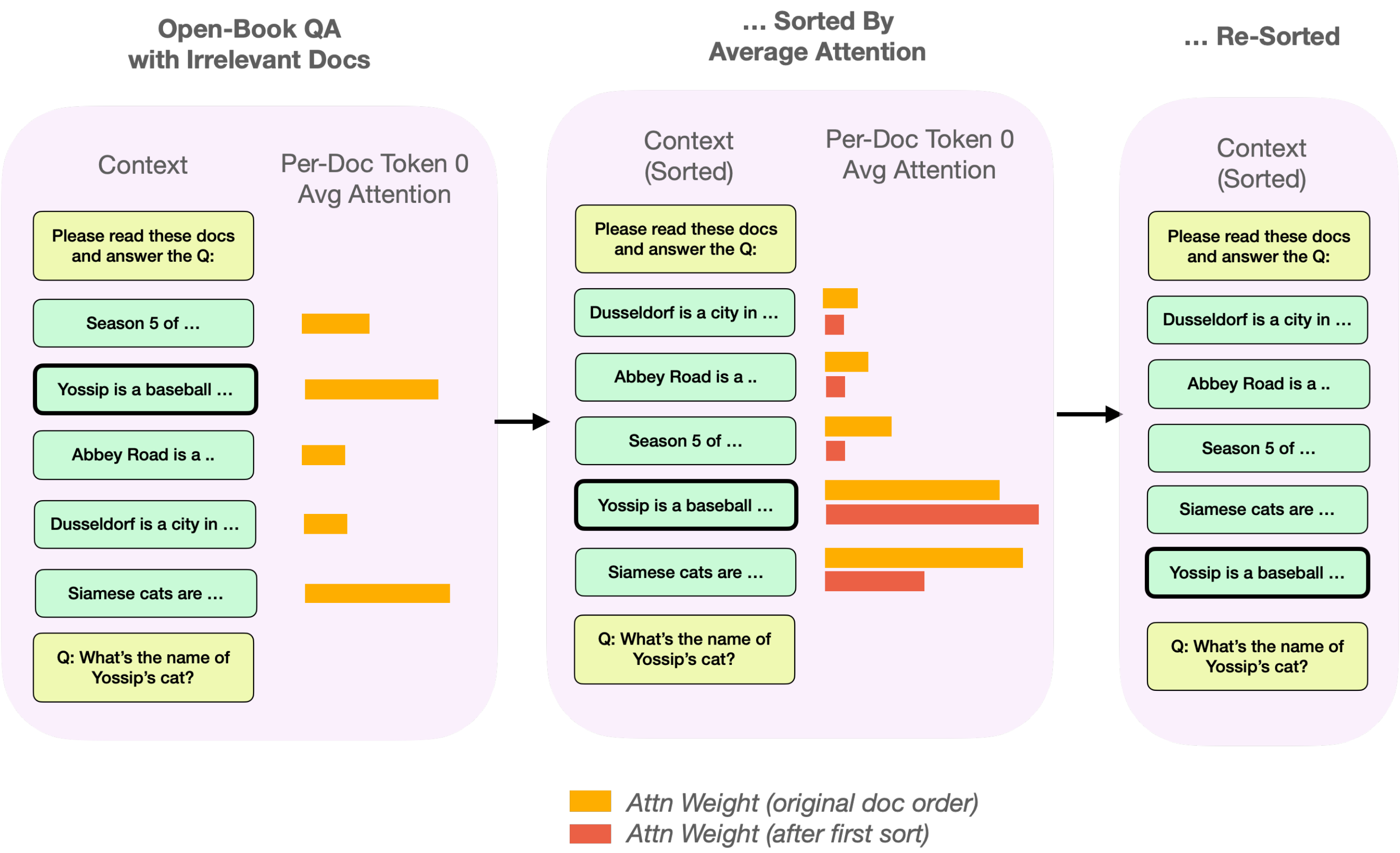}
    \caption{An illustration of the attention sorting procedure. Average per-document attention is computed for the first generated response token, and then documents are sorted in context with the highest attention at the end. After k rounds of this sorting procedure, the response is generated.}
    \label{fig:smooth_exp}
\end{figure}

Our findings suggest a simple intervention: compute average document attention weights for the first step of decoding, and then re-sort the documents, putting those with highest attention weights last. We illustrate this method in Figure \ref{fig:smooth_exp}.

\begin{figure}[h!]
    \centering
    \includegraphics[scale=.75]{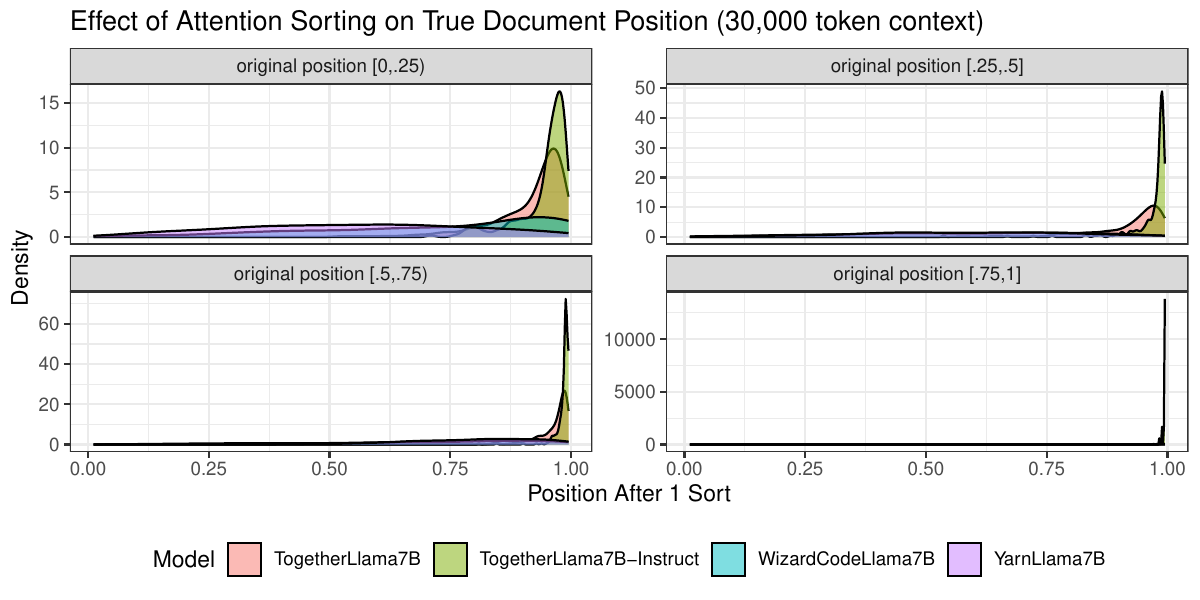}
    \caption{The effect of attention sorting on true document position. Each panel shows the distribution of document positions after one round of attention sorting for documents that began in a certain quartile of context position. In general, the true document moves towards the end of the context, however, when the true document is towards the beginning of the context it does not move all the way to the end in one shot. Attention sorting works to some extent for all models, but those that that were trained on long-context QA (i.e. the Together models) attend more effectively across the context and thus do better.}
    \label{fig:attn_sort1}
\end{figure}

Figure \ref{fig:attn_sort1} shows the results of doing attention sorting in long contexts. On average, relevant documents move closer to the end of the document list with each successive re-sort. We see that on average, a re-sorting iteration moves the relevant document closer to the end of context but for documents far away from the end, one single sort may leave them in the middle of the context.

This last fact motivates us to compute attention weights and re-sort documents in context multiple times\footnote{Alternatively, it might be possible to correct for a per-position estimated attention bias, which would avoid multiple steps of re-sorting, though it is not clear how general such a computation would be across tasks or even across multiple questions within the same QA.}.

\begin{figure}[h!]
    \centering
    \includegraphics[scale=.65]{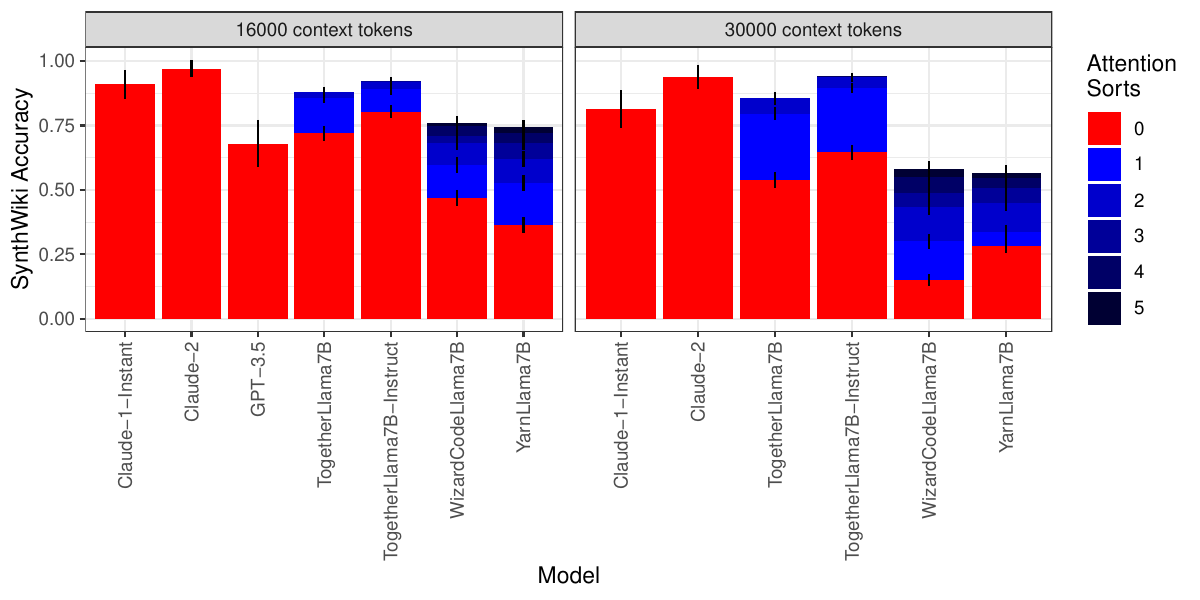}
    \caption{The effect of attention sorting on SynthWiki. Attention sorting increases small model performance. For the TogetherLlama-Instruct model, re-sorting recovers most of the performance degradation from long context and matches the performance of Claude-2.}
    \label{fig:main_fig}
\end{figure}

In the bottom panel of Figure \ref{fig:main_fig} we see that attention sorting improves accuracy on \wikitask. For TogetherLlama, after $2$ iterations we no longer gain much increase in QA performance. For other models we continue to see accuracy increases all the way up to $5$ attention sorts, because the recency bias is so strong on these models that documents in the middle of the context take multiple sorts to get moved to the end. 

\section{Some Other Stuff We Tried}
We now discuss some other results that we found in our explorations which may be of interest to specialist readers.

First, we can break down the attention score weighting by layer. We see that the differential attention to true vs. distractor documents happens mostly in the middle layers. It is not clear what to make of this fact. We did not find that restricting our attention sorting calculation to middle layers improved performance noticeably. We report it here for the interested reader.

\begin{figure}[h!]
    \centering
    \includegraphics[scale=.75]{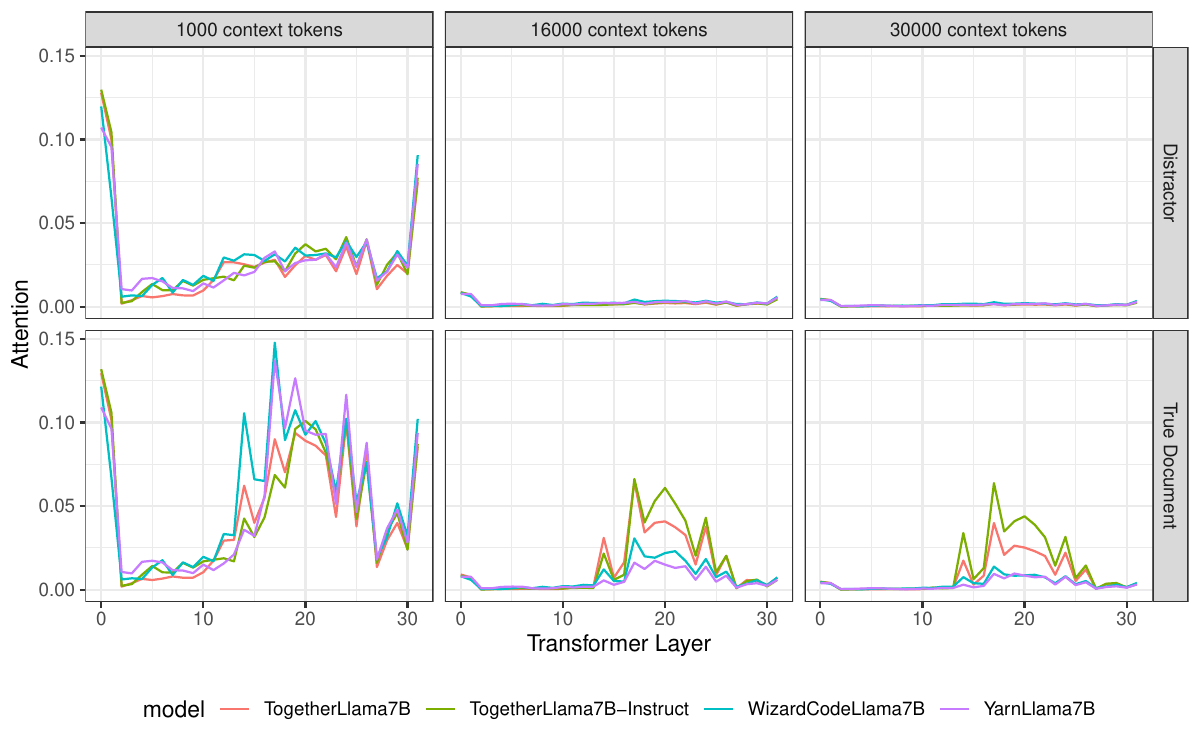}
    \caption{Average attention weight by document position for different context lengths. The attention score is for the first decode token, and is averaged across all heads and all tokens within a document. Attention score differentiation between true and distractor documents happens mostly in middle layers in all three Llama2-based models. (Attention scores do not sum to 1 because the model may also attend to non-document tokens such as the question.) }
    \label{fig:attn_layers}
\end{figure}

We hypothesized that performance degradation comes not just from the positional bias (which attention sorting addresses), but also a general dilution of attention. Since softmax attention weights are positive and sum to $1$, having many irrelevant documents monotonically increases noise that enters the representation. So we tried several methods of explicit attention truncation. 

We tried truncating the context to the top-K documents by average attention weight and found that this improved performance somewhat with one re-sort but after multiple re-sorts, truncation no longer provided much benefit. However, this could present an efficiency optimization, because re-sorting can be performed with a single step of decoding, and then the full decode can be performed with a truncated context.

We also tried to directly truncate the \textit{attention} distribution using top-k or nucleus sampling \citep{holtzman2019curious}. Unfortunately without any fine-tuning in this regime this led to serious degradation of model performance.




\section{Related Work}

\textbf{Long-Context Models} Self-attention is quadratic in the sequence length so transformer variants that can operate in a compute-efficient manner on long contexts is an active area of research, see \citep{tay2020long} for a useful overview. Perhaps the most important innovation enabling long context models was FlashAttention, which implements standard attention but uses tiling and recomputation to avoid quadratic memory consumption and more efficiently utilize GPU memory bandwidth\citep{dao2022flashattention}.

Recently the open source community has found some tricks that allow extension of context length for open source models like Llama2 (pre-trained with a 4K context length) at the fine-tuning stage, leading to the release of long-context versions of these models \citep{chen2023extending,peng2023yarn}.

Recent work has shown that long-context models do not use the information in the middle of their context efficiently \citep{liu2023lost}. Anthropic has also published guidelines for effective prompting of long-context models, which highlights that important information should be placed at the end of context\citep{anthropic2023prompt}. They suggest using a model scratchpad to pull important information of context to the end, which is quite similar to attention sorting (although operating directly in the space of text).

\textbf{Retrieval-Augmented Generation}
Retrieval-augmented generation is a framework that augments parametric language models with the ability to retrieve information from a non-parametric text database \citep{lewis2020retrieval}. Typically, a retrieval model selects passages from the database and places them in the context of the generator LM. The RAG framework affords several advantages: access to a huge memory without increasing model flops or parameter count, the ability to locate the provenance of a model response, and the ability to adapt on the fly to different settings (e.g. different codebases) without model retraining.

RAG is typically formulated as the combination of a retrieval model that selects a set of documents to be placed in the context of a generator model, that generates a response. Since the performance of the RAG system is bounded by the recall of the retriever and thus the number of documents returned, larger context windows can naturally improve the performance of these systems.

Prior work have demonstrated superior language modeling and question answering performance with fewer parameters by utilizing a huge database of text (more than seen at training time) \citep{lewis2020retrieval,guu2020retrieval,karpukhin2020dense}. Retrieval-augmented LMs typically perform especially well on open-domain QA, where an explicit knowledge DB (e.g. wikipedia) can drastically improve performance \citep{lewis2020retrieval}. 

Retrieval-like methods can conversely be used to improve long-context generation. The Memorizing Transformer improves long-context language modeling by performing an approximate kNN lookup into a long context history based on an attention-like mechanism, rather than directly attending to the full history \citep{wu2022memorizing}. This retrieval mechanism provides superior long context performance to architectures that compress the long context into a fixed-size representation such as \citep{dai2019transformer}. In Memorizing Transformer, retrieval is formulated as a compute-efficient way to condition on a long context, but our work suggests that the attention sparsity induced by retrieval may also improve the performance of the model.

\textbf{Attention-Based Salience}
Attention weights have long been used to visualize the salience of different parts of a model's input on its predictions, including the original papers proposing the attention mechanism \citep{bahdanau2014neural} (see \citep{belinkov2019analysis} Sec. 3 for a survey).

In the interpretability literature there is some disagreement about how well-suited attention weights are as explanatory variables 
\citep{vashishth2019attention,serrano2019attention,bastings2020elephant}. Attention weights have also been used as a saliency signal for various purposes across other modalities including vision and neural models for biology \citep{carion2020end,xu2015show,rao2020transformer}. 

While most retrieval systems are based on semantic similarity between document and query, some prior work has suggested using attention distillation for  a retrieval signal \citep{izacard2020distilling}. In that work, synthetic labels for query-document pairs are generated from a language model based on the attention between query and document when placed in context, and these labels are used to train a retriever model. Our attention sorting is closely related to this idea.

\section{Conclusion and Limitations}
In this work, we investigated a potential source of quality degradation on long-context tasks and demonstrated a cheap inference-time trick - attention sorting - that helps address it.

Attention sorting has several limitations. Many long-context tasks do not map neatly to a set of permutable documents; in these settings one might still want to move certain passages to the end of context but it's less clear how to do so while maintaining the prompt's original semantics. In addition, in context augmented tasks such as code generation the granularity (file, function, etc...) of the permutable elements is a priori unclear.

In addition, real long-context retrieval tasks typically have relevant rather than just random distractors. Such `hard' distractors may actually be picked up by attention sorting. If we consider a RAG task where some documents may contain `false facts' or if distractor documents are designed adversarially then a pure-attention based retriever/re-ranker will not work as well. Combining attention sorting with other retrieval objectives (eg. perplexity distillation, semantic similarity, high quality filtering, factuality) is an interesting area for future work.

Nevertheless, the attention sorting trick is an interesting example of how a model can be leveraged to modify its own context in a way that improves its ability to perform a task. While prior work such as chain of thought \citep{wei2022chain} and scratchpads \citep{nye2021show} work in the space of language, we use the attention weights as a signal. The idea of using a model to iteratively refine its own context over multiple passes may also be useful for other situations.

While attention-sorting could be applied to an `off the shelf' LLM, the fact that it works better with a model specifically trained on long-context QA suggests that a more principled solution that will likely yield better gains in real RAG tasks is a joint fine-tuning of the LLM model with its retriever. Thus, we suggest that an even more important direction for future work is studying how to achieve this harmonizing efficiently.


\section{Acknowledgements}
Experiments were performed on hardware provided by Sutter Hill Ventures. We thank Arthur Szlam for helpful feedback on an early version of this manuscript.

\bibliographystyle{plainnat} 
\bibliography{references}

\newpage

\appendix

\section{\wikitask Dataset Construction}
\label{app:synthwiki_construction}

The \wikitask dataset was constructed as follows:

First, a set of origins and jobs were constructed by querying GPT-3.5 with the following prompts:

\begin{lstlisting}
    Give me a list of 50 origins for people (example: Canadian, Texan, European, Monagasque), output it as a comma separated list. Output the list only, nothing else.
\end{lstlisting}

\begin{lstlisting}
    Give me a list of 50 jobs that people can become famous for (example: programmer, entrepreneur, taxi driver, basketball player), output it as a comma separated list. Output the list only, nothing else.
\end{lstlisting}

For each entry, a random origin and job was selected and a name for the person was generated with the following prompt to GPT-3.5:

\begin{lstlisting}
    I am writing a novel, help me come up with a name for a famous {an_origin} {a_job}. Do not output the name of someone who already exists. Output only the name, no explanation.
\end{lstlisting}

Finally, for $500$ unique (name, origin, job) pair, GPT-4 was queried with the prompt:

\begin{lstlisting}
    Please write a one paragraph wikipedia article for a famous {origin} {job} named {name}.
    Make sure the article contains information that can answer the following questions:
    {question1}
    {question2}
    Output the article only, no extraneous explanation.
\end{lstlisting}

The two questions were selected randomly from the following list:

\begin{lstlisting}
    "In which city was {person} born?",
    "What year was {person} born?",
    "Where did {person} go to college?",
    "What is the name of {person}'s spouse?",
    "What is the name of the first company {person} worked at?",
    "What is the company {person} founded called?",
    "What is the title of the film {person} directed?",
    "Who is {person}'s idol?",
    "What is the name of {person}'s pet?",
    "What is {person}'s favorite color?",
    "Where did {person} go to high school?",
    "What is the name of {person}'s best friend?",
    "What is the title of {person}'s favorite movie?",
    "In what year did {person} get married?",
    "What is the title of {person}'s favorite book?",
    "What is the name of {person}'s first child?",
    "What is the name of {person}'s favorite sports team?",
    "In which country was {person} born?",
    "What was the title of {person}'s PhD thesis?",
    "What sport does {person} play?",
\end{lstlisting}

The two questions used in the construction prompt are also those used in the QA task.

\section{\wikitask Prompt Construction}
\label{app:synthwiki_prompt}

The prompt for the Wizard model is given explicitly on the model repository. Note that the original prompt contains `Response' instead of `Answer'. We found that the models had better accuracy when we used `Answer' instead of `Response', so these are the numbers we report.

\begin{tcolorbox}
\begin{lstlisting}
Below is an instruction that describes a task. Write a response that appropriately completes the request.

### Instruction
Here is some information you will use to answer a question. Some of the information may be irrelevant.

### Information
DOCUMENT: {document}

DOCUMENT: {document}

DOCUMENT: {document}

...

### Question
{question}

Please return only the answer to the question. Answer concisely.

### Answer
\end{lstlisting}
\end{tcolorbox}

The prompt format for Together-Llama-Instruct is also given explicitly on the model repository.
\begin{tcolorbox}
\begin{lstlisting}
[INST]
Here is some information you will use to answer a question. Some of the information may be irrelevant.

### Information
DOCUMENT: {document}

DOCUMENT: {document}

DOCUMENT: {document}

...

### Question
{question}

Please return only the answer to the question. Answer concisely.
[/INST]
\end{lstlisting}
\end{tcolorbox}

\end{document}